# Hard-Aware Point-to-Set Deep Metric for Person Re-identification

Rui Yu[1], Zhiyong Dou[1], Song Bai[1], Zhaoxiang Zhang[2], Yongchao Xu[1(✉)], and Xiang Bai[1(✉)]

[1] Huazhong University of Science and Technology, Wuhan, China
{yurui.thu, songbai.site}@gmail.com , {zydou,yongchaoxu,xbai}@hust.edu.cn
[2] Institute of Automation, Chinese Academy of Sciences, Beijing, China
zhaoxiang.zhang@ia.ac.cn

**Abstract.** Person re-identification (re-ID) is a highly challenging task due to large variations of pose, viewpoint, illumination, and occlusion. Deep metric learning provides a satisfactory solution to person re-ID by training a deep network under supervision of metric loss, *e.g.*, triplet loss. However, the performance of deep metric learning is greatly limited by traditional sampling methods. To solve this problem, we propose a Hard-Aware Point-to-Set (HAP2S) loss with a soft hard-mining scheme. Based on the point-to-set triplet loss framework, the HAP2S loss adaptively assigns greater weights to harder samples. Several advantageous properties are observed when compared with other state-of-the-art loss functions: 1) *Accuracy*: HAP2S loss consistently achieves higher re-ID accuracies than other alternatives on three large-scale benchmark datasets; 2) *Robustness*: HAP2S loss is more robust to outliers than other losses; 3) *Flexibility*: HAP2S loss does not rely on a specific weight function, *i.e.*, different instantiations of HAP2S loss are equally effective. 4) *Generality*: In addition to person re-ID, we apply the proposed method to generic deep metric learning benchmarks including CUB-200-2011 and Cars196, and also achieve state-of-the-art results.

**Keywords:** Person re-identification · Deep metric learning · Triplet loss

## 1 Introduction

Person re-identification (re-ID) [47] is a task of searching designated individuals from a large amount of pedestrian images captured by non-overlapping cameras. It attracts extensive attention owing to its significance in video surveillance. The large variations of pose, viewpoint, illumination and occlusion make person re-ID a very challenging problem.

Thanks to the great success of deep learning in computer vision, along with the release of more and more large-scale person re-ID datasets [16,27,46], deep convolutional neural network (CNN) becomes a mainstream person re-ID method for learning discriminative feature representations, which have been proven superior to traditional hand-crafted features.

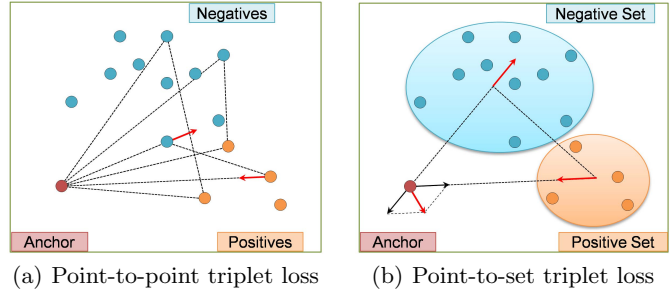

(a) Point-to-point triplet loss    (b) Point-to-set triplet loss

**Fig. 1.** (a) Traditional point-to-point (P2P) triplet loss suffers from the sampling problem. Two of three selected triplets are useless for training. (b) The proposed loss is a point-to-set (P2S) triplet loss.

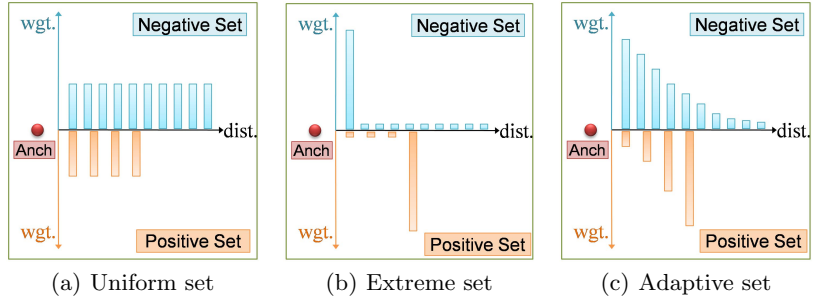

(a) Uniform set    (b) Extreme set    (c) Adaptive set

**Fig. 2.** Three types of weight distribution for computing the P2S distance. (a) Uniform set. All samples share equal weights. (b) Extreme set. Only the hardest sample is allotted a non-zero weight. (c) Adaptive set. Each sample is allotted a weight according to its "difficulty level". Note that for positive set, the hard samples are the ones far from anchor; while for negative set, the hard samples are near to anchor.

Since person re-ID can be regarded as a retrieval task in the testing phase, deep metric learning provides an effective methodology for training deep person re-ID models. Accordingly, metric loss function becomes a research hotspot in person re-ID. Several variants of metric loss have been widely applied to person re-ID, including contrastive loss [37], triplet loss [8,10], and quadruplet loss [4].

It is worth noting that the performance of metric loss is significantly influenced by the sampling method. Taking triplet loss as an example, it chooses three samples (anchor, positive, and negative) to constitute a triplet and generally aims to constrain anchor-to-positive distance less than anchor-to-negative distance. Since a training set with $N$ samples can generate $\mathcal{O}(N^3)$ triplets, it becomes infeasible to use all possible triplets even for a dataset of moderate size (*e.g.*, $N = 10^5$). Meanwhile, in practice, a large proportion of selected triplets have already satisfied the constraint and are useless. As shown in Fig. 1(a), the red, orange and blue points represent the anchor, positives and negatives, respectively. We choose three triplets and connect the points within each triplet to form three triangles. Only the triplet marked with red arrows is effective in training, while the other two triplets already satisfy the constraint and become junk. Thus, it is crucial for metric loss to select hard samples, which are able to violate the constraint and produce gradients with sufficiently large magnitude.

There are two types of hard sample mining approaches for triplet network: offline mining and online mining. Considering the offline methods are time consuming [10,1], we focus on the online mining manner. Unlike other online methods, we do not design an empirical rule for selecting triplets in a mini-batch. Instead, we propose a novel metric loss, namely Hard-Aware Point-to-Set (HAP2S) loss, which involves an adaptive hard mining mechanism.

First, we generalize the point-to-point (P2P) triplet loss to the point-to-set (P2S) triplet loss, as shown in Fig. 1(b). Given an anchor point, all the positive points and negative points in a mini-batch constitute the positive set and negative set, respectively. The P2S triplet loss constrains the distances from the anchor to the positive/negative set in a similar way as P2P triplet loss.

The key issue of P2S triplet loss is how to define the point-to-set distance. For that, we formulate the anchor-to-set distance by integrating the anchor-to-point distances and consider the contribution (*i.e.*, weight) of each P2P distance to the P2S distance. If all points have the same weight, as the uniform set illustrated in Fig. 2(a), the hard and easy samples are treated equally. The uniform set violates the principle of hard sample mining, and we find it almost impossible to conduct effective training in practice. Conversely, Hermans *et al.* [10] suggest choosing the hardest positive and hardest negative in a mini-batch to form the triplet, discarding the other samples. From the perspective of P2S distance, the method in [10] just assigns zero weight to all points other than the hardest samples (see the extreme set in Fig. 2(b)). However, this solution has two weaknesses: 1) It ignores other hard samples in the set which can also contribute to optimizing the network in the training, and thus may lead to a suboptimal solution. 2) It is easily influenced by a few error-labeled samples, *i.e.*, outliers, because the outliers usually act as the hardest samples and result in an undesired backpropagation.

To overcome these weaknesses, the proposed HAP2S loss introduces a soft hard-mining scheme rather than the traditional "select-or-unselect" manner. The basic idea is the harder samples adaptively gain more weights, as the adaptive set illustrated in Fig. 2(c).

Our main contribution is the HAP2S loss for person re-ID. To demonstrate its effectiveness, we evaluate the performance on three large-scale benchmark datasets including Market-1501 [46], CUHK03 [16], and DukeMTMC-reID [27,49]. Through experiments, HAP2S loss exhibits several advantages over other state-of-the-art deep metrics. 1) *Accuracy*: HAP2S loss consistently yields higher re-ID accuracies than other alternatives. Simply based on the widely-used ResNet-50 [9] model, HAP2S loss can achieve state-of-the-art performances on the three person re-ID datasets. 2) *Robustness*: Through the experiments training with outliers, HAP2S loss is empirically more robust to outliers than other losses.

3) *Flexibility*: HAP2S loss does not depend on a specific weight function. Two instantiations of HAP2S loss achieve similar re-ID accuracies, implying that the effectiveness of HAP2S loss derives from the essential idea of P2S metric and hard-aware weighting. 4) *Generality*: HAP2S loss is also effective for generic deep metric learning tasks, and yields state-of-the-art results on two popular benchmarks including CUB-200-2011 [38] and Cars196 [13].

## 2   Related Works

**Categorization of deep learning methods for re-ID.** Based on testing manner, we categorize deep learning methods for person re-ID into two types: 1) binary classification of image-pair representation; 2) computing distance between single-image representations. The first type is known as verification network [16], which is usually trained under supervision of binary softmax loss, *i.e.*, verification loss. The second type aims to learn a discriminative single-image representation for certain distance metric, *e.g.*, L2-norm. There are two popular categories of loss functions for training single-image representation. One category is multi-class classification loss, *e.g.*, softmax loss (*a.k.a.* cross-entropy loss), which is also called identification loss in person re-ID [47]. The other category is metric loss, *e.g.*, triplet loss [8], which defines a metric among samples to compute the loss. The proposed HAP2S loss belongs to metric loss.

**Metric losses for re-ID.** A variety of metric losses have been used for training deep re-ID models. Varior *et al.* [37] compute the contrastive loss in a siamese architecture. In [8], the standard triplet loss is applied to person re-ID. Cheng *et al.* [7] add an additional positive-pair constraint to the original triplet loss. Based on the triplet loss, Chen *et al.* [4] propose a quadruplet loss, which further forces the intra-class distance less than the inter-class distance between two other classes. All the metric losses mentioned above are point-to-point losses, the performance of which are greatly influenced by the sampling schemes. Unlike them, our HAP2S loss, as a point-to-set loss, includes all samples in a mini-batch and implicitly applies a soft hard-sampling scheme when computing loss. It is worth mentioning that Zhou *et al.* [52] also propose a point-to-set loss for re-ID. The P2S in [52] is composed of a pairwise term, a triplet term, and a regularizer term. They assign equal weight to each sample in the marginal set for the triplet term, while our HAP2S loss adaptively allocates larger weights to the harder samples and thus significantly outperforms the P2S loss in [52].

**Hard sample mining.** As previously mentioned, the mining of hard samples plays an essential role in the performance of deep metric learning for person re-ID. Ahmed *et al.* [1] iteratively select the samples that perform worst on current model as the hard negative samples for fine-tuning. Shi *et al.* [30] suggest choosing the moderate positive samples by comparing with the hardest negative sample in a mini-batch. Chen *et al.* [4] propose a margin-based online hard negative mining method customized for their quadruplet loss. Hermans *et al.* [10] select the hardest positive and hardest negative of each anchor in a mini-batch to compute the triplet loss. All these methods select hard samples in a "select-or-unselect" manner. In contrast, our HAP2S loss introduces a more robust hard mining strategy by adaptively assigning larger weights to the harder samples.

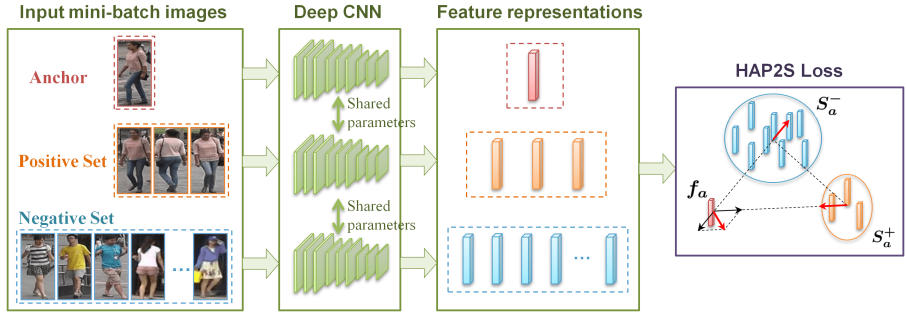

**Fig. 3.** The pipeline of the deep network using the proposed HAP2S loss

## 3 Proposed Method

### 3.1 Overview

The objective of deep metric learning is to learn a deep neural network that maps an image $x$ to a corresponding feature representation $f_\Theta(x)$, which is suited to a predefined metric. The parameters of the network, *i.e.*, the weights and biases, are included in $\Theta$. As for person re-ID, we can extract the features of probe and gallery images through a well-trained deep model and then compute the distances between the features to obtain a ranking list. The role of metric loss is to provide a discriminative metric for supervising the network training.

The proposed pipeline is depicted in Fig. 3. We adopt a pre-trained CNN model as the backbone network, which transfers each pedestrian image into an intermediate feature embedding. The backbone network employed in this work is ResNet-50 [9] model, which consists of five down-sampling blocks and one global average pooling layer. The backbone network is followed by two fully connected (FC) layers with 1024 and 128 neurons, respectively. After backbone and two FC layers, the network extracts the output features to compute HAP2S loss in a mini-batch. The Euclidean distance is employed as the point-to-point (P2P) metric. Details about the proposed method is given in the following.

### 3.2 Revisit Triplet Loss

Triplet loss [40,28] is one of the most representative metric losses. Given a training mini-batch $\boldsymbol{X} = \{\boldsymbol{x}_i\}_{i=1}^{N_s}$ with labels $\{y_i\}_{i=1}^{N_s}$, we select a triplet $\{\boldsymbol{x}_a, \boldsymbol{x}_p, \boldsymbol{x}_n\}$ where the *anchor* $\boldsymbol{x}_a$ and the *positive* $\boldsymbol{x}_p$ are two images from the same person, while the *negative* $\boldsymbol{x}_n$ is an image from another person. The corresponding features are $f_\Theta(\boldsymbol{x}_a)$, $f_\Theta(\boldsymbol{x}_p)$, and $f_\Theta(\boldsymbol{x}_n)$. To simplify the notation, we use $\boldsymbol{f}_a$ to replace $f_\Theta(\boldsymbol{x}_a)$, and so forth. Despite of several variants [7,25], the most common expression of triplet loss is as follows

$$\mathcal{L}_{trp} = \frac{1}{N_t} \sum_{\substack{y_p = y_a \\ y_n \neq y_a}} \left[ d(\boldsymbol{f}_a, \boldsymbol{f}_p) - d(\boldsymbol{f}_a, \boldsymbol{f}_n) + m \right]_+, \tag{1}$$

where $[\cdot]_+ = \max(\cdot, 0)$, $N_t$ represents the number of all possible triplets in the mini-batch, and $d$ is a predefined distance metric. It can be seen from Eq.(1) that triplet loss aims to force the distance between an intra-class pair less than an inter-class pair by at least a margin $m$. While training a CNN with triplet loss, many of the possible triplets would easily satisfy the constraint

$$d(\boldsymbol{f}_a, \boldsymbol{f}_p) + m < d(\boldsymbol{f}_a, \boldsymbol{f}_n). \tag{2}$$

It makes the selected triplet equal to 0, *i.e.*, useless for training. Thus, the hard sample mining is critical to triplet loss. Hermans *et al.* [10] present a variant of triplet loss with a simple yet powerful hard-mining scheme, defined as

$$\mathcal{L}_{trpBH} = \frac{1}{N_s} \sum_{a=1}^{N_s} \left[ \max_{y_p = y_a} d(\boldsymbol{f}_a, \boldsymbol{f}_p) - \min_{y_n \neq y_a} d(\boldsymbol{f}_a, \boldsymbol{f}_n) + m \right]_+, \tag{3}$$

where the hardest positive and hardest negative for each anchor in a mini-batch (Batch Hard) are selected to constitute a triplet. Based on this variant, they reported state-of-the-art results on two large-scale datasets.

### 3.3 Hard-Aware P2S Loss

Triplet loss is a type of P2P loss, since it only includes distances between points. Though interesting results can be obtained with the P2P triplet loss using a simple hard-mining scheme [10], as discussed in Sec. 1, such a simple hard-mining scheme may lead to two problems: 1) it excludes the contributions of other hard samples in the gradient descent training; 2) it is vulnerable to the outliers which usually serve as the hardest samples, causing undesired backpropagation. These two problems reveal that the simple solution of hard mining is not robust enough.

In this work, we generalize the P2P triplet loss to point-to-set (P2S) triplet loss. Given an anchor with label $y_a$, let $\boldsymbol{S}_a^+ = \{\boldsymbol{f}_p | y_p = y_a\}$ denote the positive set which contains all positive points in the mini-batch and similarly $\boldsymbol{S}_a^- = \{\boldsymbol{f}_n | y_n \neq y_a\}$ be the negative set. The P2S triplet loss is defined as

$$\mathcal{L}_{P2S} = \frac{1}{N_s} \sum_{a=1}^{N_s} \left[ D\left(\boldsymbol{f}_a, \boldsymbol{S}_a^+\right) - D\left(\boldsymbol{f}_a, \boldsymbol{S}_a^-\right) + m \right]_+, \tag{4}$$

where $D$ represents the P2S distance. The P2S triplet loss is a more generic form, which can be transferred to the P2P triplet loss in Eq.(3) if the P2S distance is

defined as

$$\begin{cases} D\left(\boldsymbol{f}_a, \boldsymbol{S}_a^+\right) = \max_{\boldsymbol{f}_i \in \boldsymbol{S}_a^+} d\left(\boldsymbol{f}_a, \boldsymbol{f}_i\right) \\ D\left(\boldsymbol{f}_a, \boldsymbol{S}_a^-\right) = \min_{\boldsymbol{f}_j \in \boldsymbol{S}_a^-} d\left(\boldsymbol{f}_a, \boldsymbol{f}_j\right) \end{cases}. \quad (5)$$

From the perspective of P2S loss, the triplet loss in Eq.(3) only selects the hardest sample to represent the whole set.

To solve the problems of P2P triplet loss described above, we present a hard-aware P2S (HAP2S) loss with an adaptive hard mining scheme. The HAP2S loss has the same form as Eq.(4). The key of HAP2S loss is to assign different weights to the points in each set by computing the P2S distance as

$$\begin{cases} D\left(\boldsymbol{f}_a, \boldsymbol{S}_a^+\right) = \dfrac{\sum\limits_{\boldsymbol{f}_i \in \boldsymbol{S}_a^+} d(\boldsymbol{f}_a, \boldsymbol{f}_i) w_i^+}{\sum\limits_{\boldsymbol{f}_i \in \boldsymbol{S}_a^+} w_i^+} \\ D\left(\boldsymbol{f}_a, \boldsymbol{S}_a^-\right) = \dfrac{\sum\limits_{\boldsymbol{f}_j \in \boldsymbol{S}_a^-} d(\boldsymbol{f}_a, \boldsymbol{f}_j) w_j^-}{\sum\limits_{\boldsymbol{f}_j \in \boldsymbol{S}_a^-} w_j^-} \end{cases}, \quad (6)$$

where $w_i^+$ and $w_j^-$ represent the weights of the elements $\boldsymbol{f}_i$ and $\boldsymbol{f}_j$ in the positive and negative set respectively. As discussed in Sec. 1, an effective hard mining strategy should assign higher weights to the hard samples in a set. Considering the metric loss, the "difficulty level" of a sample lies in the distance from the anchor to it. Accordingly, for the positive set, the remote points to the anchor are the hard ones and deserve higher weights. On the contrary, for the negative set, the nearest point to the anchor is the hardest. To this end, we introduce two weighting schemes to the proposed HAP2S loss.

**i) Exponential weighting.** The first weighting scheme is exponential weighting. The weights of the elements in each set are defined as

$$\begin{cases} w_i^+ = \exp\left(\dfrac{d(\boldsymbol{f}_a, \boldsymbol{f}_i)}{\sigma}\right) & \text{if } \boldsymbol{f}_i \in \boldsymbol{S}_a^+ \\ w_j^- = \exp\left(-\dfrac{d(\boldsymbol{f}_a, \boldsymbol{f}_j)}{\sigma}\right) & \text{if } \boldsymbol{f}_j \in \boldsymbol{S}_a^- \end{cases}, \quad (7)$$

where $\sigma > 0$ is a coefficient for adjusting the weight distribution. In this way, the weight of each sample exponentially adapts to its "difficulty level". The complete formula of HAP2S loss $\mathcal{L}_{HAP2S}$ is composed of Eq.(4), Eq.(6) and Eq.(7).

**ii) Polynomial weighting.** Instead of exponential weighting, we can define an alternate HAP2S loss by assigning weights to the elements in each set via a univariate polynomial function with real coefficients, as

$$\begin{cases} w_i^+ = \left(d\left(\boldsymbol{f}_a, \boldsymbol{f}_i\right) + 1\right)^{\alpha} & \text{if } \boldsymbol{f}_i \in \boldsymbol{S}_a^+ \\ w_j^- = \left(d\left(\boldsymbol{f}_a, \boldsymbol{f}_j\right) + 1\right)^{-2\alpha} & \text{if } \boldsymbol{f}_j \in \boldsymbol{S}_a^- \end{cases}, \quad (8)$$

where $\alpha > 0$ is also a coefficient for adjusting the weight distribution. The weighting scheme of Eq. (8) is similar to that of Eq. (7) by assigning greater weights to harder samples. This instantiation of $\mathcal{L}_{HAP2S}$ consists of Eq.(4), Eq.(6) and Eq.(8). In order distinguish the two instantiations, we denote the former one as **HAP2S_E** and the latter one as **HAP2S_P**.

To demonstrate why HAP2S loss outperforms other alternatives, we analyze the gradient to show how the loss optimizes the network parameters. Due to space limitation, the detailed analyses are given in the supplementary material.

### 3.4 Multi-loss Training

Since various losses establish different optimization objectives, joint supervision of different losses usually helps to train better deep re-ID models. For example, McLaughlin *et al.* [23] adopt softmax loss and contrastive loss to train a recurrent neural network; Chen *et al.* [5] optimize a multi-task deep network jointly by verification loss, triplet loss, and contrastive loss.

We notice that the metric loss (*e.g.*, HAP2S loss) does not fully utilize the annotations provided by the training set. It only verifies the labels of two samples, but ignores the specific class ID. In contrast, the classification loss (*e.g.*, softmax loss) exactly uses the multi-class labels as the supervision information. Based on this observation, we can combine the proposed HAP2S loss with softmax loss to further boost re-ID performance. The details of multi-loss training (including network, algorithm, and experiments) are given in the supplementary material.

## 4 Experiments

### 4.1 Datasets and Evaluation Protocols

We evaluate the proposed method on three challenging large-scale benchmark datasets including Market-1501 [46], CUHK03 [16], and DukeMTMC-reID [27,49].

**Market-1501.** This dataset comprises 32,668 labeled images of 1,501 identities captured by six cameras. Within the dataset, 12,936 images of 751 identities are used for training, while the rest are used for testing. Among the testing data, fixed 3,368 images constitute the probe set. The testing set also contains 2,793 distractor images, which makes this dataset very challenging.

**CUHK03.** This dataset consists of 14,096 images of 1,467 identities captured by six cameras. Each identity shows in two camera views and has 4.8 images on average in one view. The dataset provides two types of data. One is a set of manually labeled bounding boxes of pedestrians, while the other set contains automatically detected bounding boxes by the DPM detector. We conduct experiments on both "labeled" and "detected" datasets.

**DukeMTMC-reID.** This dataset contains 36,411 images of 1,812 identities, which are manually cropped from multi-camera tracking dataset DukeMTMC [27]. The images are captured by eight cameras, and 1,404 identities appear in more than one camera. Following [49], the 1,404 identities are divided into two halves, with 702 identities for training and the others for testing.

**Table 1.** Comparison with other losses based on pre-trained ResNet-50 model. Note that test-phase data augmentations are *not* applied throughout the experiments.

| Loss | Market1501 | | CUHK03 (labeled) | | | CUHK03 (detected) | | | DukeMTMC | |
|---|---|---|---|---|---|---|---|---|---|---|
| | mAP | r=1 | r=1 | r=5 | r=10 | r=1 | r=5 | r=10 | mAP | r=1 |
| Softmax [47] | 58.16 | 79.25 | 72.81 | 95.09 | 97.89 | 71.61 | 92.20 | 95.64 | 49.33 | 71.01 |
| Triplet [8] | 53.40 | 70.84 | 78.11 | 97.17 | 98.15 | 75.71 | 94.55 | 97.07 | 48.16 | 66.02 |
| Improved Triplet[7] | 55.10 | 71.56 | 80.56 | 97.69 | 98.48 | 76.10 | 95.43 | 97.36 | 49.63 | 66.92 |
| P2S [52] | 54.30 | 70.99 | 77.82 | 97.67 | 98.40 | 75.94 | 96.01 | 97.64 | 49.74 | 66.88 |
| OIM [41] | 60.48 | 81.26 | 77.88 | 94.86 | 97.54 | 75.09 | 93.41 | 95.84 | 51.32 | 71.99 |
| Quadruplet [4] | 64.88 | 81.47 | 85.17 | 97.64 | 98.52 | 84.13 | 97.70 | 98.55 | 54.29 | 73.47 |
| Hard Triplet [10] | 67.22 | 82.13 | 87.65 | 98.46 | 99.37 | 86.34 | 97.28 | 98.65 | 57.08 | 74.37 |
| HAP2S_E | **69.76** | 84.20 | 90.22 | 98.76 | 99.37 | 88.13 | 97.67 | 98.69 | 59.58 | **76.08** |
| HAP2S_P | 69.43 | **84.59** | **90.39** | **99.54** | **99.90** | **88.90** | **98.44** | **99.09** | **60.64** | 75.94 |

**Evaluation protocols.** We adopt the standard evaluation protocols. For Market-1501 and DukeMTMC-reID, since the data are divided fixedly, we directly evaluate the cumulative matching characteristics (CMC) and mean average precision (mAP), and report the average results of two independent trials. For CUHK03, 100 identities are selected for testing and the rest are used for training. The CUHK03 provides 20 different train/test splits, so we report the average CMC on the 20 trails. All experiments are by default under the single query setting. We also report multiple query evaluation results for Market-1501.

### 4.2 Implementation Details

We implement the deep model on PyTorch framework. The network is trained with Adam [12] under supervision of HAP2S loss. The learning rate is $4 \times 10^{-4}$ at the first 100 epochs and gradually decay to $4 \times 10^{-7}$ at the 150th epoch.

All images are first resized to $256 \times 128$. Standard random crop and horizontal flipping are adopted for data augmentation in training phase. Following [8,10], we select a fixed number of images for each person to form a mini-batch. In our experiments, eight images from each of 32 persons are randomly chosen as a 256-size mini-batch. We set the parameters of HAP2S loss with margin $m = 2.5$, and weight coefficient $\sigma = 0.5$ (*resp.* $\alpha = 10$) for HAP2S_E (*resp.* HAP2S_P). It costs less than an hour to train the model on two GTX TITAN X 12GB GPUs.

In testing phase, we do not apply any data augmentations (*e.g.*, five crops and flips [10]) on account of efficiency. We extract the intermediate feature given by the backbone network for each image and produce the ranking results according to Euclidean metric. Note that the results of multi-loss training (Sec. 3.4) are not reported below, but given in the supplementary material.

### 4.3 Comparison with Other Losses

We compare HAP2S loss with state-of-the-art losses reported in recent person re-ID works, including softmax loss [47], triplet loss [8], improved triplet loss [7], P2S loss [52], OIM loss [41], quadruplet loss [4], and hard triplet loss [10]. We separately evaluate the performance of each loss with the pre-trained ResNet-50 model on the three datasets. To be fair in comparison, we apply the same mini-batch configuration and tune the parameters to optimum for each loss. The hard mining approaches presented by original papers are also reproduced.

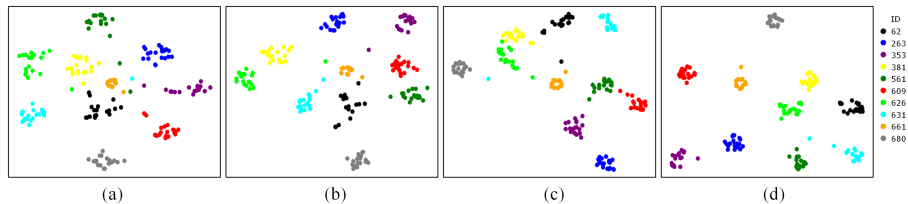

**Fig. 4.** Visualization of deeply-learned features by (a) softmax loss [47], (b) triplet loss [8], (c) hard triplet loss [10], (d) HAP2S loss using exponential weighting. We randomly select 10 identities from the testing set of Market-1501. The points with different colors denote features from different identities. (Best viewed in color)

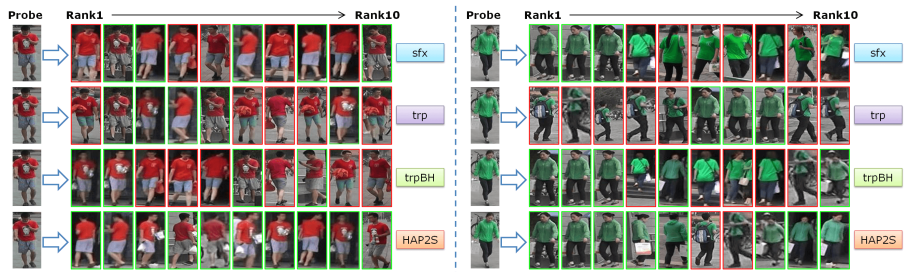

**Fig. 5.** Examples of some retrieval results on Market-1501. Each row contains the top-10 rank images retrieved by the corresponding method (*sfx* - softmax loss [47]; *trp* - triplet loss [8]; *trpBH* - hard triplet loss [10]). The correct and false matches are enclosed in green and red boxes, respectively. (Best viewed in color)

**HAP2S loss consistently outperforms other alternatives.** The experimental results are presented in Table 1. As can be seen, the identification losses (softmax [47] and OIM [41]) achieve favorable re-ID accuracies, while the metric losses (*e.g.*, quadruplet [4] and hard triplet [10]) can realize higher performances with certain hard mining strategies. The proposed HAP2S loss using either exponential weighting or polynomial weighting outperforms all other competitors on the three datasets. On Market-1501 and DukeMTMC-reID, the performance gaps between HAP2S and other losses are consistently more than +2.2% in mAP. As for CUHK03, HAP2S loss also outperforms other alternatives with a noticeable improvement in rank-1 accuracy on both labeled and detected datasets. It is worth noting that the results of HAP2S_P are on par with that of HAP2S_E. Thus, HAP2S loss does not depend on a specific weight function. More generally, it is expected that any weighting functions with similar properties as Eq. (7) or Eq.(8) would produce other effective instantiations of HAP2S loss.

Table 2. Comparison with state-of-the-art methods. (Best results are highlighted)

| Method | Market-1501 | | | | CUHK03 | | | | | | DukeMTMC | |
|---|---|---|---|---|---|---|---|---|---|---|---|---|
| | single query | | multi query | | labeled | | | detected | | | | |
| | mAP | r=1 | mAP | r=1 | r=1 | r=5 | r=10 | r=1 | r=5 | r=10 | mAP | r=1 |
| OL-MANS [51] | - | 60.67 | - | 74.00 | 54.7 | 86.5 | 93.9 | 45.0 | 76.0 | 83.5 | - | - |
| DNS [43] | 35.68 | 61.02 | 46.03 | 71.56 | 62.6 | 90.1 | 94.8 | 54.7 | 84.8 | 94.8 | - | - |
| Gated SCNN[36] | 39.55 | 65.88 | 48.45 | 76.04 | - | - | - | 68.1 | 88.1 | 94.6 | - | - |
| P2S [52] | 44.27 | 70.72 | 55.73 | 85.78 | - | - | - | - | - | - | - | - |
| MTDnet [5] | - | - | - | - | 74.7 | 96.0 | 97.5 | - | - | - | - | - |
| CRAFT-MFA[6] | 42.30 | 68.70 | 50.30 | 77.00 | - | - | - | 84.3 | 97.1 | 98.3 | - | - |
| ACRN [29] | 62.60 | 83.61 | - | - | - | - | - | 62.6 | 89.7 | 94.7 | 51.96 | 72.58 |
| CADL [19] | 47.11 | 73.84 | 55.58 | 80.85 | - | - | - | - | - | - | - | - |
| MSCAN [15] | 57.53 | 80.31 | 66.70 | 86.79 | 74.2 | 94.3 | 97.5 | 68.0 | 91.0 | 95.4 | - | - |
| Quadruplet [4] | - | - | - | - | 75.5 | 95.2 | 99.2 | - | - | - | - | - |
| PAN [48] | 63.35 | 82.81 | 71.72 | 88.18 | - | - | - | - | - | - | 51.51 | 71.59 |
| Re-ranking [50] | 63.63 | 77.11 | - | - | 61.6 | - | - | 58.5 | - | - | - | - |
| APR [20] | 64.67 | 84.29 | - | - | - | - | - | - | - | - | 51.88 | 70.69 |
| SSM [2] | 68.80 | 82.21 | 76.18 | 88.18 | 76.6 | 94.6 | 98.0 | 72.7 | 92.4 | 96.1 | - | - |
| MuDeep [26] | - | - | - | - | 76.9 | 96.1 | 98.4 | 75.6 | 94.4 | 97.5 | - | - |
| SVDNet [35] | 62.10 | 82.30 | - | - | - | - | - | 81.8 | 95.2 | 97.2 | 56.80 | 76.70 |
| JLML [17] | 65.50 | 85.10 | 74.50 | 89.70 | 83.2 | 98.0 | 99.4 | 80.6 | 96.9 | 98.7 | - | - |
| Part-Aligned[45] | 63.40 | 81.00 | - | - | 85.4 | 97.6 | 99.4 | 81.6 | 97.3 | 98.4 | - | - |
| Spindle [44] | - | 76.90 | - | - | 88.5 | 97.8 | 98.6 | - | - | - | - | - |
| PDC [34] | 63.41 | 84.14 | - | - | 88.7 | 98.6 | 99.2 | 78.3 | 94.8 | 97.2 | - | - |
| LSRO [49] | 66.07 | 83.97 | 76.10 | 88.42 | - | - | - | 84.6 | 97.6 | 98.9 | 47.13 | 67.68 |
| MLFN [3] | 74.30 | 90.00 | 82.40 | 92.30 | - | - | - | 82.8 | - | - | 62.80 | **81.00** |
| HA-CNN [18] | **75.70** | **91.20** | **82.80** | **93.80** | - | - | - | - | - | - | **63.80** | 80.50 |
| HAP2S_E | 69.76 | 84.20 | 76.50 | 88.69 | 90.2 | 98.8 | 99.4 | 88.1 | 97.7 | 98.7 | 59.58 | 76.08 |
| HAP2S_P | 69.43 | 84.59 | 76.75 | 90.20 | **90.4** | **99.5** | **99.9** | **88.9** | **98.4** | **99.1** | 60.64 | 75.94 |

**Visualization analysis.** The t-SNE [21] tool is adopted to visualize the feature embeddings learned by the losses. We randomly choose 10 identities and 20 images for each identity from the testing set of Market-1051. The visualization results of the features are plotted in Fig. 4. As it shows, HAP2S loss achieves larger inter-class variances and smaller intra-class variances than other losses. In addition, we show several example re-ID results on Market-1501 in Fig. 5. HAP2S loss can find more correct matches than other losses in top ranks.

In sum, based on the quantitative and qualitative results, the superiority of HAP2S loss is not only proved by the better retrieval performance (see Table 1), but also by the superior clustering quality (see Fig. 4).

### 4.4  Comparison with State-of-the-arts

We compare in Table 2 the proposed method with the state-of-the-art methods on the three datasets described in Sec. 4.1.

**Comparison on Market-1501.** For the sake of testing efficiency, we apply neither test-phase augmentation nor post-ranking. Despite this, the proposed method still outperforms most of state-of-the-arts under both single query (SQ) and multiple query (MQ) settings. When compared with a recently reported multi-loss model JLML [17], the proposed method achieves an improvement of about +4% mAP for SQ. It is also worth mentioning that the performance of the proposed method can be further boosted by re-ranking tools. For example, when applying the re-ranking approach [50], the proposed method can achieve 79.91% mAP and 85.72% rank-1 accuracy for SQ.

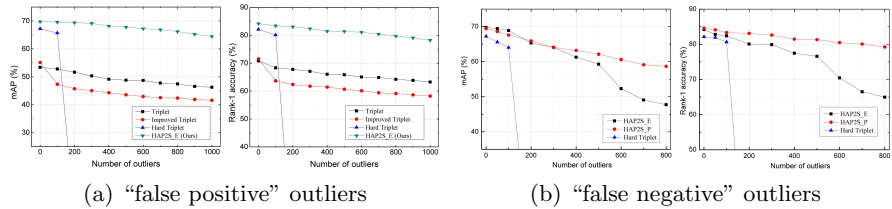

(a) "false positive" outliers        (b) "false negative" outliers

**Fig. 6.** Re-ID results on Market-1501 with different numbers of outliers in training set.

**Comparison on CUHK03.** The proposed method yields the best rank-1 accuracies on both labeled and detected datasets. For labeled dataset, the pose-driven deep convolutional (PDC) model [34] reports previous best performance of 88.7% in rank-1 accuracy. The proposed method outperforms PDC [34] on the labeled dataset by about +1.5% rank-1 accuracy, and increases the performance gap with PDC by +9.8% rank-1 accuracy on the detected dataset. Therefore, the proposed method exhibits greater robustness than the body-part-based method in the automatically detected scenario, which is closer to practical applications.

**Comparison on DukeMTMC-reID.** The state-of-the-art performances are achieved by MLFN [3] and HA-CNN [18]. In [3], a novel multi-level factorization net is proposed to learn latent discriminative factors without manual annotation. In [18], a well-designed CNN can learn discriminative features by soft pixel attention and hard regional attention. The loss function in [3] and [18] is softmax loss. By contrast, we only use pre-trained ResNet-50 network. It can be expected that HAP2S loss would outperform current state-of-the-arts by using better backbone networks such as HA-CNN [18].

### 4.5   Robustness to outliers

In order to assess the robustness of different losses to outliers, we conduct experiments with outliers. Specifically, we randomly select a certain number of images from CUHK03 and add them to the training set of Market-1501 as outliers. Each outlier is randomly labeled with an ID number from the 751 identities of Market-1501 training set. Then, we use the new training set to train a deep network and conduct the standard evaluation on the testing set of Market-1501.

Several variants of triplet loss are evaluated in this experimental setting. The re-ID performances with varying number of outliers are depicted in Fig. 6(a). The proposed HAP2S loss is least affected by outliers. Even when 1,000 outliers are present, HAP2S_E loss still achieves a high performance of 64.48% mAP.

**Table 3.** Re-ID results on Market-1501 with different $\sigma$ (HAP2S_E) or $\alpha$ (HAP2S_P)

| $\sigma$ | $\sigma\to 0$[10] | $\sigma=0.25$ | $\sigma=0.5$ | $\sigma=0.75$ | $\sigma=1$ | $\sigma=1.25$ | $\sigma=1.5$ | $\sigma=2$ | $\sigma=4$ | $\sigma=8$ | $\sigma\to\infty$ |
|---|---|---|---|---|---|---|---|---|---|---|---|
| mAP | 67.22 | 68.86 | **69.76** | 69.25 | 68.21 | 67.88 | 67.23 | 66.57 | 63.68 | 62.30 | 1.36 |
| r=1 | 82.13 | 83.28 | **84.20** | 83.76 | 82.48 | 82.27 | 81.09 | 79.87 | 78.15 | 76.22 | 3.30 |

| $\alpha$ | $\alpha=0$ | $\alpha=4$ | $\alpha=5$ | $\alpha=6$ | $\alpha=7$ | $\alpha=8$ | $\alpha=9$ | $\alpha=10$ | $\alpha=11$ | $\alpha=12$ | $\alpha\to\infty$[10] |
|---|---|---|---|---|---|---|---|---|---|---|---|
| mAP | 3.11 | 59.07 | 64.75 | 67.46 | 68.48 | 69.29 | 69.39 | **69.43** | 69.20 | 68.94 | 67.22 |
| r=1 | 7.27 | 76.07 | 80.26 | 82.93 | 83.76 | 84.03 | 84.31 | **84.59** | 83.64 | 83.21 | 82.13 |

Traditional triplet loss [8] is more robust to outliers than the improved triplet loss [7]. In line with previous analyses, hard triplet loss [10] is very sensitive to outliers. When the number of outliers is larger than 200, the model training with hard triplet loss is prone to collapse due to multiple outliers in a mini-batch.

Besides the "false positive" outliers, we also report the re-ID performances with "false negative" outliers. In particular, we randomly assign false ID labels to varying number of images in original Market-1501 training set. The evaluation results depicted in Fig. 6(b) also verifies the robustness of HAP2S loss.

### 4.6 Parameter Analysis

For either the exponential weighting in Eq.(7) or the polynomial weighting in Eq.(8), a coefficient ($\sigma$ or $\alpha$) is introduced to adjust the weight distribution. Here, we empirically analyze the impact of the coefficient. According to Eq.(7) and Eq.(8), HAP2S loss is equivalent to hard triplet loss [10] when $\sigma\to 0$ or $\alpha\to\infty$. Conversely, HAP2S loss turns into uniform weighting when $\sigma\to\infty$ or $\alpha=0$.

We evaluate the re-ID performance of HAP2S loss on Market-1501 dataset under different parameters. As seen in Table 3, HAP2S loss can yield both high accuracy and training stability when $\sigma \leq 1.5$ or $\alpha \geq 6$. The uniform weighting ($\sigma\to\infty$ or $\alpha=0$) tends to produce no loss and fail in training.

### 4.7 Generic Deep Metric Learning

**Datasets and evaluation metrics.** The proposed HAP2S deep metric has demonstrated its superiority on person re-ID task in previous experiments. To further assess the effectiveness of HAP2S loss, we evaluate it on two popular deep metric learning benchmarks: CUB-200-2011 [38] and Cars196 [13]. We follow the same training/testing split and standard protocol described in [33,31]. The *CUB-200-2011* dataset [38] consists of 11,788 bird images of 200 species, where the first 100 species (5,864 images) are used for training and the others for testing. The *Cars196* dataset [13] contains 16,185 car images of 196 classes. We use the first 98 classes (8,054 images) for training and the rest for testing. The performances are measured by two standard metrics. The normalized mutual information (NMI) [22] is used for evaluating the clustering quality, while the Recall@K metric [11] serves to measure the retrieval performance.

**Comparison with state-of-the-arts.** We adopt the same network architecture as for person Re-ID task (Fig. 3) for training and testing. As the proposed

Table 4. Comparison with state-of-the-art methods of generic deep metric learning

| Method | CUB-200-2011 | | | | | Cars196 | | | | |
|---|---|---|---|---|---|---|---|---|---|---|
| | NMI | R@1 | R@2 | R@4 | R@8 | NMI | R@1 | R@2 | R@4 | R@8 |
| Lifted Struct [33] | 56.50 | 43.57 | 56.55 | 68.59 | 79.63 | 56.88 | 52.98 | 65.70 | 76.01 | 84.27 |
| N-pairs [31] | 57.24 | 45.37 | 58.41 | 69.51 | 79.49 | 57.79 | 53.90 | 66.76 | 77.75 | 86.35 |
| Clustering [32] | 59.23 | 48.18 | 61.44 | 71.83 | 81.92 | 59.04 | 58.11 | 70.64 | 80.27 | 87.81 |
| Proxy NCA [24] | 59.53 | 49.21 | 61.90 | 67.90 | 72.40 | **64.90** | 73.22 | 82.42 | 86.36 | 88.68 |
| Smart Mining [14] | 59.90 | 49.78 | 62.34 | 74.05 | 83.31 | 59.50 | 64.65 | 76.20 | 84.23 | 90.19 |
| HDC [42] | - | 53.60 | 65.70 | 77.00 | 85.60 | - | 73.70 | 83.20 | 89.50 | 93.80 |
| Angular Loss [39] | 61.10 | 54.70 | 66.30 | 76.00 | 83.90 | 63.20 | 71.40 | 81.40 | 87.50 | 92.10 |
| HAP2S_E | **63.42** | **56.08** | **68.43** | **79.20** | **86.93** | 63.05 | **74.09** | **83.45** | **89.85** | **94.13** |

HAP2S_E and HAP2S_P loss performs similarly, for simplicity, we only compare HAP2S_E with state-of-the-art methods, as shown in Table 4. The proposed HAP2S loss outperforms the other methods by about +2% in terms of both NMI and Recall@K scores on the CUB-200-2011 dataset. For the Cars196 dataset, HAP2S loss achieves the best Recall@K scores in all ranks, while reaching comparable NMI with the state-of-the-art methods.

## 5  Conclusions

The selection of training samples is a crucial problem in deep metric learning for person re-ID. In this paper, we propose a novel loss function, Hard-Aware Point-to-Set (HAP2S) loss, to solve the sampling problem in a robust manner. Unlike traditional solutions, HAP2S loss does not focus on how to select samples, but to distribute different weights to the samples. Based on the P2S triplet loss framework, HAP2S loss adaptively assigns greater weights to harder samples. We conduct extensive experiments on three large-scale person re-ID benchmarks. Benefiting from the soft hard-mining scheme, HAP2S loss achieves state-of-the-art re-ID accuracies on the three datasets. Besides, HAP2S loss performs more robust than other alternatives when some outliers are present in the training set. Moreover, HAP2S loss is also able to yield state-of-the-art performances on generic deep metric learning benchmarks. In this work, we mainly target on loss function and adopt the off-the-shelf ResNet-50 network. It can be expected that the performance of the proposed HAP2S loss would be further boosted by bespoke re-ID networks [3,18].

**Acknowledgements.** This work was supported by National Key R&D Program of China No. 2018YFB1004600, NSFC 61703171, and NSFC 61573160, to Dr. Xiang Bai by the National Program for Support of Top-notch Young Professionals and the Program for HUST Academic Frontier Youth Team. We would also like to thank the reviewers for their helpful comments.

# Supplementary Material of Hard-Aware Point-to-Set Deep Metric for Person Re-identification

Rui Yu[1], Zhiyong Dou[1], Song Bai[1], Zhaoxiang Zhang[2], Yongchao Xu[1(✉)], and Xiang Bai[1(✉)]

[1] Huazhong University of Science and Technology, Wuhan, China
{yurui.thu, songbai.site}@gmail.com , {zydou,yongchaoxu,xbai}@hust.edu.cn
[2] Institute of Automation, Chinese Academy of Sciences, Beijing, China
zhaoxiang.zhang@ia.ac.cn

We propose a Hard-Aware Point-to-Set (HAP2S) loss for person re-ID in the main paper, and show that it consistently achieves higher re-ID accuracies than other losses on three large-scale datasets. In this supplementary material, we first provide more experimental results to further demonstrate the superiority of HAP2S loss. Then, we analyze how HAP2S loss optimizes the network parameters to show why HAP2S loss outperforms other alternatives. Finally, a multi-loss training method is introduced to further improve the performance of HAP2S loss.

## 1 Further Experimental Analysis

### 1.1 Robustness to Distance Measure

It is reported in [5] that hard triplet training is prone to collapse when using squared Euclidean. We observed the same phenomenon in our experiments on Market-1501 as shown in Table 1. Besides, we noticed that smooth $\ell_1$ distance can also make the hard triplet training collapsing. In contrast, the two variants of HAP2S loss consistently achieve high performances under all the three distance measures. The robustness to distance measure shows another advantage of HAP2S loss over hard triplet loss [5].

### 1.2 Impact of Batch Size

Figure 1 shows the re-ID performances of hard triplet [5] and HAP2S loss on Market-1501 across different batch sizes. With the increase of batch size, the mAP and rank-1 accuracy rise steadily for all the three losses. It can be interpreted as that larger batch size can provide more hard samples for effective training. Although achieving better performance, larger batch size will demand more computing resources. It is also noted that the two instantiations of HAP2S loss consistently outperform hard triplet loss [5] across all tested batch sizes.

Table 1. Re-ID results on Market-1501 with different distance measure

| Loss | Euclidean | | Squared Euclidean | | Smooth $\ell_1$ | |
|---|---|---|---|---|---|---|
| | mAP | r=1 | mAP | r=1 | mAP | r=1 |
| Hard Triplet [5] | 67.22 | 82.13 | 0.58 | 0.86 | 0.21 | 0.06 |
| HAP2S_E | **69.76** | 84.20 | **68.74** | **83.79** | **68.45** | **83.97** |
| HAP2S_P | 69.43 | **84.59** | 66.64 | 82.63 | 67.40 | 83.37 |

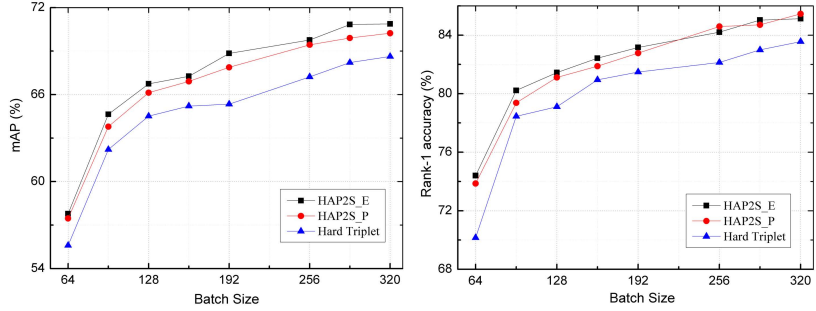

**Fig. 1.** Re-ID results on Market-1501 across different batch sizes

## 2   Gradient Analysis of HAP2S Loss

### 2.1   Detailed Gradient Analysis

We propose two variants of HAP2S loss, *i.e.*, HAP2S_E and HAP2S_P, in the main paper. Here we derive the gradients of HAP2S_E as an example. The analysis for HAP2S_P is similar. As introduced in the main paper, the main equations of HAP2S_E for an anchor are given as below

$$\mathcal{L}_{HAP2S} = \left[ D\left(\boldsymbol{f}_a, \boldsymbol{S}_a^+\right) - D\left(\boldsymbol{f}_a, \boldsymbol{S}_a^-\right) + m \right]_+, \tag{1}$$

$$\begin{cases} D\left(\boldsymbol{f}_a, \boldsymbol{S}_a^+\right) = \dfrac{\sum\limits_{\boldsymbol{f}_i \in \boldsymbol{S}_a^+} d(\boldsymbol{f}_a, \boldsymbol{f}_i) w_i^+}{\sum\limits_{\boldsymbol{f}_i \in \boldsymbol{S}_a^+} w_i^+} \\ D\left(\boldsymbol{f}_a, \boldsymbol{S}_a^-\right) = \dfrac{\sum\limits_{\boldsymbol{f}_j \in \boldsymbol{S}_a^-} d(\boldsymbol{f}_a, \boldsymbol{f}_j) w_j^-}{\sum\limits_{\boldsymbol{f}_j \in \boldsymbol{S}_a^-} w_j^-} \end{cases}, \tag{2}$$

$$\begin{cases} w_i^+ = \exp\left(\dfrac{d(\boldsymbol{f}_a, \boldsymbol{f}_i)}{\sigma}\right) & \text{if } \boldsymbol{f}_i \in \boldsymbol{S}_a^+ \\ w_j^- = \exp\left(-\dfrac{d(\boldsymbol{f}_a, \boldsymbol{f}_j)}{\sigma}\right) & \text{if } \boldsymbol{f}_j \in \boldsymbol{S}_a^- \end{cases}. \tag{3}$$

In order to analyze how HAP2S loss optimizes the parameters $\boldsymbol{\Theta}$ of the network, we adopt the following notations to facilitate demonstration:

$$\begin{cases} d_i^+ = d\left(\boldsymbol{f}_a, \boldsymbol{f}_i\right) & \text{for } \boldsymbol{f}_i \in \boldsymbol{S}_a^+ \\ d_j^- = d\left(\boldsymbol{f}_a, \boldsymbol{f}_j\right) & \text{for } \boldsymbol{f}_j \in \boldsymbol{S}_a^- \end{cases} \quad (4)$$

and $\mathcal{T} = D\left(\boldsymbol{f}_a, \boldsymbol{S}_a^+\right) - D\left(\boldsymbol{f}_a, \boldsymbol{S}_a^-\right) + m$. So, the derivatives of HAP2S loss can be computed as

$$\frac{\partial \mathcal{L}_{HAP2S}}{\partial \boldsymbol{\Theta}} = \begin{cases} \frac{\partial D(\boldsymbol{f}_a, \boldsymbol{S}_a^+) - \partial D(\boldsymbol{f}_a, \boldsymbol{S}_a^-)}{\partial \boldsymbol{\Theta}} & \text{if } \mathcal{T} > 0 \\ 0 & \text{if } \mathcal{T} \leqslant 0 \end{cases}. \quad (5)$$

By the definition of $D\left(\boldsymbol{f}_a, \boldsymbol{S}_a^+\right)$ in Eq.(2), we can compute its gradient as

$$\frac{\partial D(\boldsymbol{f}_a, \boldsymbol{S}_a^+)}{\partial \boldsymbol{\Theta}} = \left[\left(\sum_i w_i^+\right)\left(\sum_i w_i^+ \frac{\partial d_i^+}{\partial \boldsymbol{\Theta}} + d_i^+ \frac{\partial w_i^+}{\partial \boldsymbol{\Theta}}\right) - \left(\sum_i d_i^+ w_i^+\right)\left(\sum_i \frac{\partial w_i^+}{\partial \boldsymbol{\Theta}}\right)\right] \bigg/ \left(\sum_i w_i^+\right)^2 \quad (6)$$

According to Eq.(3), the gradient of $w_i^+$ is

$$\frac{\partial w_i^+}{\partial \boldsymbol{\Theta}} = \frac{1}{\sigma} e^{\frac{d_i^+}{\sigma}} \frac{\partial d_i^+}{\partial \boldsymbol{\Theta}}. \quad (7)$$

Then we can substitute $w_i^+$ in Eq.(6) and derive that

$$\frac{\partial D(\boldsymbol{f}_a, \boldsymbol{S}_a^+)}{\partial \boldsymbol{\Theta}} = \left[\sum_{i \neq j}\left(1 + \frac{d_i^+ - d_j^+}{\sigma}\right) e^{\frac{d_i^+ + d_j^+}{\sigma}} \frac{\partial d_i^+}{\partial \boldsymbol{\Theta}} + \sum_i e^{\frac{2d_i^+}{\sigma}} \frac{\partial d_i^+}{\partial \boldsymbol{\Theta}}\right] \bigg/ \left(\sum_i e^{\frac{d_i^+}{\sigma}}\right)^2 \quad (8)$$

Similarly, we have

$$\frac{\partial D(\boldsymbol{f}_a, \boldsymbol{S}_a^-)}{\partial \boldsymbol{\Theta}} = -\left[\sum_{i \neq j}\left(1 + \frac{d_j^- - d_i^-}{\sigma}\right) e^{-\frac{d_i^- + d_j^-}{\sigma}} \frac{\partial d_i^-}{\partial \boldsymbol{\Theta}} + \sum_i e^{-\frac{2d_i^-}{\sigma}} \frac{\partial d_i^-}{\partial \boldsymbol{\Theta}}\right] \bigg/ \left(\sum_i e^{-\frac{d_i^-}{\sigma}}\right)^2 \quad (9)$$

It can be observed from Eq.(8) that all samples in the positive set participate in the gradient computation, and the samples with larger distances to the anchor have greater weights in the derivative of $D\left(\boldsymbol{f}_a, \boldsymbol{S}_a^+\right)$. Conversely, in the negative set, the samples with smaller distances obtain greater weights in the derivative of $D\left(\boldsymbol{f}_a, \boldsymbol{S}_a^-\right)$ as shown in Eq.(9). In sum, the harder samples contribute more to the derivative of HAP2S loss. This adaptive hard mining scheme in training process makes HAP2S loss superior to other alternatives.

If squared Euclidean distance is utilized to compute the P2P distance

$$d\left(\boldsymbol{f}_i, \boldsymbol{f}_j\right) = \left\|\boldsymbol{f}_i - \boldsymbol{f}_j\right\|_2^2. \quad (10)$$

Then, we can derive the gradients of the anchor-to-point distances

$$\begin{cases} \frac{\partial d_i^+}{\partial \boldsymbol{\Theta}} = 2\left(\boldsymbol{f}_a - \boldsymbol{f}_i\right) \frac{\partial \boldsymbol{f}_a - \partial \boldsymbol{f}_i}{\partial \boldsymbol{\Theta}} & \text{for } \boldsymbol{f}_i \in \boldsymbol{S}_a^+ \\ \frac{\partial d_i^-}{\partial \boldsymbol{\Theta}} = 2\left(\boldsymbol{f}_a - \boldsymbol{f}_j\right) \frac{\partial \boldsymbol{f}_a - \partial \boldsymbol{f}_j}{\partial \boldsymbol{\Theta}} & \text{for } \boldsymbol{f}_j \in \boldsymbol{S}_a^- \end{cases} \quad (11)$$

4    R. Yu, Z. Dou, S. Bai, Z. Zhang, Y. Xu, and X. Bai

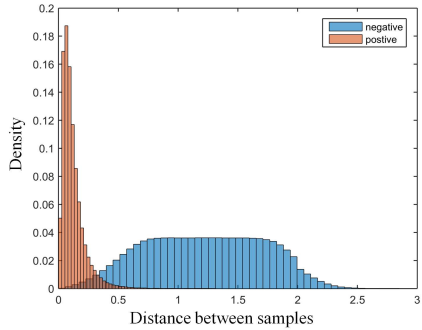 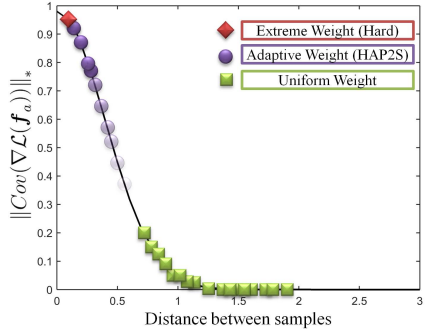

**Fig. 2.** Pairwise distance distributions for positive and negative pairs

**Fig. 3.** Sample distributions for different weighting schemes.

Since $\boldsymbol{f}_i$ and $\frac{\partial \boldsymbol{f}_i}{\partial \boldsymbol{\Theta}}$ can be obtained by the standard forward and backward propagations for each image in the mini-batch, we can compute the partial derivative of HAP2S loss according to Eq.(5), Eq.(8), Eq.(9) and Eq.(11).

### 2.2 More Intuitive Analysis

Inspired by [8], we regard our weighting scheme as a form of sampling. Then, we can analyze the gradient of HAP2S loss in a more intuitive manner. The gradient of HAP2S $w.r.t.$ the anchor $\boldsymbol{f}_a$ has the form of

$$\frac{\partial \mathcal{L}_{HAP2S}}{\partial \boldsymbol{f}_a} = \sum_{\boldsymbol{f}_p \in \boldsymbol{S}_a^+} \frac{\Delta_{ap}}{\|\Delta_{ap}\|} g(a,p) + \sum_{\boldsymbol{f}_n \in \boldsymbol{S}_a^-} \frac{\Delta_{an}}{\|\Delta_{an}\|} h(a,n), \qquad (12)$$

where $\Delta_{ai} = \boldsymbol{f}_a - \boldsymbol{f}_i$, $i \in \{p,n\}$, and $\frac{\Delta_{ai}}{\|\Delta_{ai}\|}$ determines the direction of each gradient component; $g$ and $h$ are functions involving weights. The weighting is equivalent to sampling with a defined probability as done in [8]. Following [8], we analyze the gradient components of negative samples.

First, we compute pairwise distance distributions for positive and negative pairs on Market-1501 based on a training model, as depicted in Fig.2. The uniform weighting scheme mainly selects negative samples with distances in the range of $(0.5, 2)$, which is generally larger than positive-pair distances. So the chosen negative samples are easy ones and induce no loss.

Then, Figure 3 shows the gradient variance curve with noise $z \sim \mathcal{N}(0, \sigma_1^2 I)$ and the corresponding "sample distributions" for different weighting schemes. The points with opaque colors are allotted greater weights than the ones with translucent colors. Hard triplet loss [5] always chooses the negative sample with the smallest distance (corresponding to the highest variance), so noisy gradient $\frac{\Delta_{an}+z}{\|\Delta_{an}+z\|}$ might cause collapsed training. In contrast, HAP2S loss adaptively focuses on hard examples, which can produce effective gradients while keeping stable training.

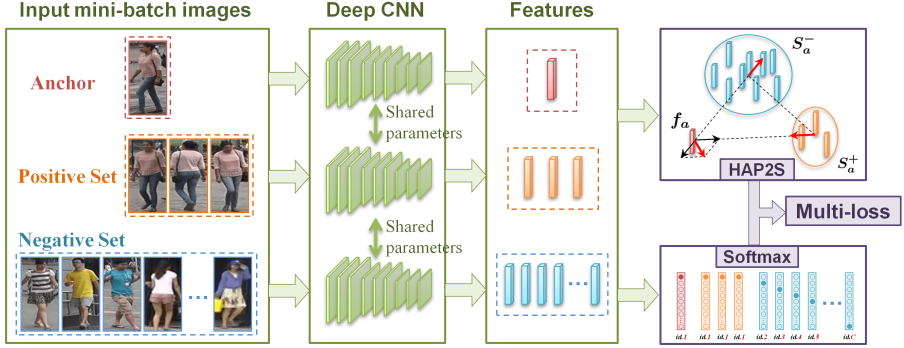

**Fig. 4.** The pipeline of the proposed multi-task network

## 3  Multi-loss Training

### 3.1  Network and Algorithm

As mentioned in the main paper, we notice that the metric loss (*e.g.*, HAP2S loss) does not fully utilize the annotations provided by the training set. It only verifies the labels of two samples, but ignores the specific class ID. In contrast, the classification loss (*e.g.*, softmax loss) exactly uses the multi-class labels as the supervision information. The most commonly used loss for classification is softmax loss (*a.k.a.* cross-entropy loss), defined as

$$\mathcal{L}_{sfx} = -\frac{1}{N_s} \sum_{i=1}^{N_s} \log \frac{e^{W_{y_i}^\mathrm{T} \boldsymbol{f}_i + b_{y_i}}}{\sum_{j=1}^{C} e^{W_j^\mathrm{T} \boldsymbol{f}_i + b_j}}, \qquad (13)$$

where $C$ is the number of classes in the training set; $W$ and $b$ are the weights and biases of the last layer of classification network, respectively. According to Eq.(13), the optimization objective of softmax loss is classifying the features into predefined categories, which is totally different from that of HAP2S loss. Based on this observation, combining HAP2S loss with softmax loss would potentially reduce the risk of overfitting and learn a more robust representation.

The HAP2S loss is combined with softmax loss by a deep multi-task network, as depicted in Fig. 4. Similar to the network in the main paper, we adopt a pre-trained ResNet-50 [4] model as the backbone network, followed by two branches. One branch is a standard $C$-class classification network which adds two fully-connected (FC) layers and computes the softmax loss. The number of neurons in the last two FC layers are 2,048 and $C$ (the number of training classes), respectively. A dropout layer is inserted between the two FC layers with dropout rate of 0.5. The other branch includes two FC layers with 1,024 and 128 neurons, and extracts the output features to compute HAP2S loss in a mini-batch. The Euclidean distance is employed as the P2P metric.

**Algorithm 1** Multi-loss learning algorithm with HAP2S loss.
---
**Input:** Training samples $\{\boldsymbol{x}_i\}$ and labels $\{y_i\}$, maximum iteration $T$, margin $m$, learning rate $\eta^t$, hyperparameter $\sigma$ and $\lambda$.
**Output:** The network parameters $\boldsymbol{\Theta}$.
1: Initialize parameters $\boldsymbol{\Theta}$ by pre-trained model;
2: **for** $t = 1, 2, \ldots, T$ **do**
3:     Compute the feature representations of the samples in a mini-batch $\{\boldsymbol{x}_i\}^t$ by the forward propagation;
4:     Compute the HAP2S loss $\mathcal{L}_{HAP2S}^t$ by Eq.(1), Eq.(2) and Eq.(3);
5:     Compute the softmax loss $\mathcal{L}_{sfx}^t$ by Eq.(13);
6:     Compute the joint loss $\mathcal{L}^t$ by Eq.(14);
7:     Compute the gradient $\frac{\partial \mathcal{L}_{HAP2S}^t}{\partial \boldsymbol{\Theta}^t}$ according to Eq.(5), Eq.(8), Eq.(9), Eq.(11);
8:     Compute the joint gradient $\frac{\partial \mathcal{L}^t}{\partial \boldsymbol{\Theta}^t} = \lambda \frac{\partial \mathcal{L}_{HAP2S}^t}{\partial \boldsymbol{\Theta}^t} + (1-\lambda) \frac{\partial \mathcal{L}_{sfx}^t}{\partial \boldsymbol{\Theta}^t}$;
9:     Update the parameters $\boldsymbol{\Theta}$ by $\boldsymbol{\Theta}^{t+1} = \boldsymbol{\Theta}^t - \eta^t \frac{\partial \mathcal{L}^t}{\partial \boldsymbol{\Theta}^t}$;
10: **end for**

We train the CNN model under joint supervision of softmax loss and HAP2S loss. The overall loss for multi-loss training is given by

$$\mathcal{L} = \lambda \mathcal{L}_{HAP2S} + (1-\lambda) \mathcal{L}_{sfx}, \qquad (14)$$

where $\lambda \in [0, 1]$ is a hyperparameter to balance the two losses. The multi-loss training process is summarized in Algorithm 1.

### 3.2   Experimental Evaluations

The multi-loss is evaluated on the same three datasets, *i.e.*, Market-1501 [10], CUHK03 [6] and DukeMTMC-reID [7,12], as in the main paper. We follow the implementations described in Sec. 4.2 of the main paper. The multi-task network is trained under the joint supervision of HAP2S and softmax loss with the hyperparameter $\lambda = 0.5$. The experimental results are reported in Table 2.

**Multi-loss can further boost performance of HAP2S loss.** As shown in Table 2, the performances of the two variants of HAP2S loss can both be further improved by multi-loss learning. On Market-1501, "HAP2S_E+Softmax" substantially outperforms HAP2S_E with the improvement of mAP +4.73% and rank-1 accuracy +5.53%. On DukeMTMC-reID, when integrated with softmax loss, the mAP and rank-1 accuracy of HAP2S_E loss rise by +3.04% and +3.00%, respectively. On CUHK03, HAP2S_E loss has already achieved relatively high performance, so the improvement brought by multi-loss learning is limited, with +0.44% in rank-1 accuracy. The performance improvements of HAP2S_P using multi-loss learning are on par with that of HAP2S_E.

**The multi-loss framework can also benefit other metric losses.** In addition to HAP2S loss, we apply the proposed multi-task network (Fig. 4) to two competitive metric losses, including quadruplet loss [1] and hard triplet

Supplementary Material of HAP2S Deep Metric for Person Re-identification 7

**Table 2.** Re-ID results of competitive individual losses and multi-losses based on the pre-trained ResNet-50 model. Note that the test-phase data augmentations are *not* applied throughout the experiments.

| Loss | Market1501 | | CUHK03 (labeled) | | | CUHK03 (detected) | | | DukeMTMC | |
|---|---|---|---|---|---|---|---|---|---|---|
| | mAP | r=1 | r=1 | r=5 | r=10 | r=1 | r=5 | r=10 | mAP | r=1 |
| Softmax [11] | 58.16 | 79.25 | 72.81 | 95.09 | 97.89 | 71.61 | 92.20 | 95.64 | 49.33 | 71.01 |
| Triplet [3] | 53.40 | 70.84 | 78.11 | 97.17 | 98.15 | 75.71 | 94.55 | 97.07 | 48.16 | 66.02 |
| Improved Triplet[2] | 55.10 | 71.56 | 80.56 | 97.69 | 98.48 | 76.10 | 95.43 | 97.36 | 49.63 | 66.92 |
| P2S [13] | 54.30 | 70.99 | 77.82 | 97.67 | 98.40 | 75.94 | 96.01 | 97.64 | 49.74 | 66.88 |
| OIM [9] | 60.48 | 81.26 | 77.88 | 94.86 | 97.54 | 75.09 | 93.41 | 95.84 | 51.32 | 71.99 |
| Quadruplet [1] | 64.88 | 81.47 | 85.17 | 97.64 | 98.52 | 84.13 | 97.70 | 98.55 | 54.29 | 73.47 |
| Hard Triplet [5] | 67.22 | 82.13 | 87.65 | 98.46 | 99.37 | 86.34 | 97.28 | 98.65 | 57.08 | 74.37 |
| HAP2S_E | 69.76 | 84.20 | 90.22 | 98.76 | 99.37 | 88.13 | 97.67 | 98.69 | 59.58 | 76.08 |
| HAP2S_P | 69.43 | 84.59 | 90.39 | **99.54** | **99.90** | 88.90 | 98.44 | 99.09 | 60.64 | 75.94 |
| Quadruplet+Softmax | 69.74 | 86.64 | 86.44 | 98.44 | 99.45 | 84.65 | 97.47 | 98.59 | 57.49 | 77.06 |
| HardTriplet+Softmax | 70.74 | 86.19 | 88.20 | 98.59 | 99.49 | 86.49 | 97.76 | 99.08 | 60.48 | 78.23 |
| HAP2S_E+Softmax | **74.49** | **89.73** | 90.66 | 99.01 | 99.59 | 88.63 | 98.19 | 99.21 | 62.62 | 79.08 |
| HAP2S_P+Softmax | 74.33 | 89.31 | **90.81** | 99.30 | 99.83 | **89.40** | **98.47** | **99.24** | **63.27** | **80.25** |

loss [5]. As seen in Table 2, when combined with softmax loss, the re-ID performances of the two metric losses can both be consistently boosted, with more than +3% improvements in mAP on Market-1501 and DukeMTMC-reID datasets. On CUHK03, multi-loss learning can also slightly improve the performances of quadruplet loss and hard triplet loss upon high re-ID accuracies.

**HAP2S still beats other competitors in multi-loss setting.** It can be observed in Table 2 that both "HAP2S_E+Softmax" and "HAP2S_P+Softmax" outperform "Quadruplet+Softmax" and "Hard Triplet +Softmax" on the three datasets. On Market-1501 and DukeMTMC-reID, the performance gaps are consistently more than +2.14% in mAP. As for CUHK03, the differences in rank-1 accuracy are also consistently more than +2%. The experiments of multi-losses once again show the superiority of HAP2S loss over other alternatives.

### 3.3 Further Comments on Multi-loss Training

We briefly analyze why multi-loss can boost re-ID performance of HAP2S loss. HAP2S loss learns discriminative geometry structure explicitly in the feature space (Fig.5(a)), while softmax loss separates the decision boundaries explicitly in the probability space (Fig.5(b)) and implicitly in the feature space by the linear or nonlinear transformations contained in the FC layers. The interaction is implicitly involved by back-propagating error differentials of both branches to the shared backbone feature during training. The branch of softmax can combat the vanishing gradient problem by providing regularization. Thus, the two-branch multi-loss network can improve the performance of individual loss.

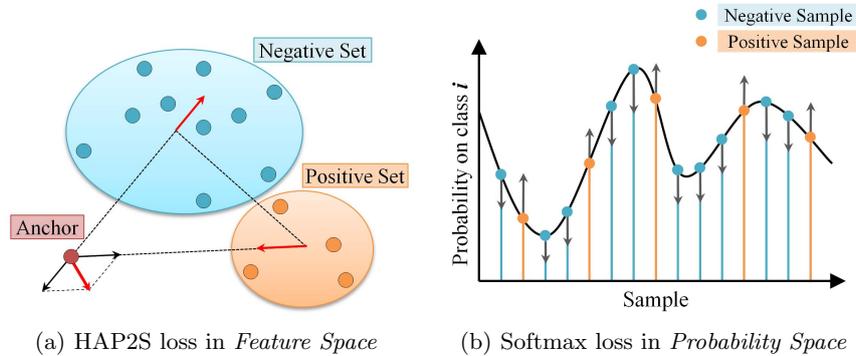

(a) HAP2S loss in *Feature Space*     (b) Softmax loss in *Probability Space*

**Fig. 5.** Illustration of optimization by (a) HAP2S and (b) softmax loss